%% file: main.tex
 \documentclass[pmlr,twocolumn,10pt]{jmlr} % W&CP article

 \usepackage{rotating} % for sideways figures and tables
\usepackage{booktabs}

\usepackage{siunitx}

\usepackage[switch]{lineno}

\newcommand{\equal}[1]{{\hypersetup{linkcolor=black}\thanks{#1}}}

\usepackage{makecell}
\usepackage{multirow}
\usepackage{pifont}
\newcommand{\cmark}{\ding{51}}
\newcommand{\xmark}{\ding{55}}
\usepackage{soul}
\usepackage{enumitem}

\jmlrvolume{297}
\jmlryear{2025}
\jmlrworkshop{Machine Learning for Health (ML4H) 2025} % W&CP title

 \title{Multimodal Cancer Modeling in the Age of Foundation Model Embeddings}

\author{%
\Name{Steven Song}\equal{These authors contributed equally} \Email{songs1@uchicago.edu}\\
\addr Center for Translational Data Science\\
\addr Department of Computer Science\\
\addr Medical Scientist Training Program\\
\addr University of Chicago, Chicago IL, USA
\AND
\Name{Morgan Borjigin-Wang}\footnotemark[1]\thanks{Work performed as a research volunteer at CTDS.} \Email{mbwang@brown.edu}\\
\addr Department of Computer Science\\
\addr Brown University, Providence RI, USA
\AND
\Name{Irene Madejski} \Email{imadejski@uchicago.edu}\\
\addr Center for Translational Data Science\\
\addr Department of Computer Science\\
\addr University of Chicago, Chicago IL, USA
\AND
\Name{Robert L. Grossman} \Email{rgrossman1@uchicago.edu}\\
\addr Center for Translational Data Science\\
\addr Department of Computer Science\\
\addr Section of Biomedical Data Science, Department of Medicine\\
\addr University of Chicago, Chicago IL, USA
}

\begin{document}

\maketitle

\begin{abstract}
The Cancer Genome Atlas (TCGA) has enabled novel discoveries and served as a large-scale reference dataset in cancer through its harmonized genomics, clinical, and imaging data. Numerous prior studies have developed bespoke deep learning models over TCGA for tasks such as cancer survival prediction. A modern paradigm in biomedical deep learning is the development of foundation models (FMs) to derive feature embeddings agnostic to a specific modeling task. Biomedical text especially has seen growing development of FMs. While TCGA contains free-text data as pathology reports, these have been historically underutilized. Here, we investigate the ability to train classical machine learning models over multimodal, zero-shot FM embeddings of cancer data. We demonstrate the ease and additive effect of multimodal fusion, outperforming unimodal models. Further, we show the benefit of including pathology report text and rigorously evaluate the effect of model-based text summarization and hallucination. Overall, we propose an embedding-centric approach to multimodal cancer modeling.
\end{abstract}


\paragraph*{Data and Code Availability}
TCGA data is publicly available through the GDC API. Extracted pathology reports from Mendely, DOI: 10.17632/hyg5xkznpx.1. Precomputed embeddings of WSIs from HuggingFace: MahmoodLab/UNI2-h-features. Our source code is available from:\linebreak
% \url{https://anonymous.4open.science/r/multimodal-cancer-survival-60E3}.
\url{https://github.com/StevenSong/multimodal-cancer-modeling}.
%%%%%%%%%%%%%%%%%%%%%%%%%%%%%%%%%%%%%%%%%%%%%%%%%%%%%%%%%%%%%%%%%%%%%%%%
%%%%%%%%%%%%%%%%%%% Deanonymize in the final version %%%%%%%%%%%%%%%%%%%
%%%%%%%%%%%%%%%%%%%%%%%%%%%%%%%%%%%%%%%%%%%%%%%%%%%%%%%%%%%%%%%%%%%%%%%%

\paragraph*{Institutional Review Board (IRB)}
This research does not require IRB approval.

\section{Introduction}

The Cancer Genome Atlas (TCGA) has been the premier cancer research resource for nearly two decades \citep{tomczak2015review}. Throughout its history, its harmonized data \citep{heath2021nci} has enabled novel discoveries through its multitudes of molecular genetic data \citep{zhang2021uniform}, histopathological images, and clinical descriptors for over 11 thousand cases across 33 cancer types. There have been numerous studies which develop deep learning methods using TCGA \citep{sartori2025comprehensive} for various tasks, however TCGA cancer survival has been particularly well studied \citep{abbasi2024survival, arjmand2022machine, liu2018integrated}.

% https://docs.google.com/spreadsheets/d/1-5Cgf73TthtpvYCWP4IO-pYFdYqldgAGFzHEFXu1iFE/edit?usp=sharing
There have been numerous studies to date that have trained bespoke models to predict patient survival from TCGA. These models have covered a wide range in the specific cancer types studied, the data modalities used as predictive features, and the model types trained. Some have leveraged histology images \citep{wulczyn2020deep, yang2025foundation} or RNA sequencing (RNAseq) \citep{ching2018cox, huang2020deep, qiu2020meta, nayshool2022surviveai} alone, while others have explored varying degrees of multimodal integration \citep{chaudhary2018deep, zhan2019correlation, hao2019interpretable, ramirez2021prediction, malik2021deep, redekar2022identification, sun2023interpretable, fan2023pancancer, hao2023cancer}. Most recent papers focus on training task-specific deep-learning models \citep{ching2018cox,hao2019interpretable,huang2020deep,wulczyn2020deep,qiu2020meta,ramirez2021prediction,malik2021deep,sun2023interpretable,fan2023pancancer,hao2023cancer,yang2025foundation} and some apply simpler machine learning models \citep{zhan2019correlation,nayshool2022surviveai,redekar2022identification}, however few explore the synergy of combining deep learning with simpler statistical, machine learning models \citep{chaudhary2018deep}.

A modern paradigm in biomedical deep learning is the development of foundation models (FMs) to derive meaningful feature embeddings \citep{bommasani2021opportunities}. These FMs are typically trained over large corpora of data using self-supervision to improve generalization of embeddings to downstream tasks. Biomedical text especially has seen growing development of FM large language models (LLMs) \citep{singhal2023large, thirunavukarasu2023large}. While many LLMs have been adapted towards the biomedical domain using research text from PubMed and PubMed Central \citep{gu2021domain, bolton2024biomedlm, labrak2024biomistral}, more specific pathology-report text FMs have also been developed as vision-language models \citep{lu2024visual, xiang2025vision}.

While TCGA contains free-text data as pathology reports, these have been historically underutilized. This may in part be due to the difficulty of working with the raw data format of these reports as scanned PDFs. With the current interest in applications of LLMs to biomedical domains, a recent effort by \citet{kefeli2024tcga} used optical character recognition to extract the text from all available PDFs of TCGA. To the best of our knowledge, only one other work has explored the application of TCGA reports towards survival modeling \citep{xiang2025vision} and differs from our approach (see Section \ref{sec:discussion}).

Here, we investigate the combination of simple machine learning models with modern FMs (Figure \ref{fig:overview}) for predicting cancer survival using multimodal data from TCGA.\\

\noindent \textbf{\ul{Our main contributions are:}}

\begin{itemize}[leftmargin=*]
    \item \textbf{Quantification of pan-cancer survival prediction using rich embeddings with small models.} We contemporize survival modeling for TCGA, the premier cancer research resource, using zero-shot, foundation model-derived embeddings combined with small models. Despite advanced deep learning methods, \ul{survival models over FM embeddings of single data modalities do not do substantially better than survival models over 5 tabular clinical features} (see Section \ref{sec:unimodal-results}). 
    \item \textbf{Simple framework for multimodal fusion of many data modalities that improves cancer survival prediction.} We introduce a modular framework to do late-fusion of unimodal models, that is extensible to variable data modalities and embedding methods. Using our approach, we demonstrate that \ul{cancer data modalities are additive and non-redundant, including tabular clinical features} (see Section \ref{sec:multimodal-results}).
    \item \textbf{Analysis on the effect of clinical text summarization and model hallucination on cancer modeling.} We develop a method for automatic pathology report summarization and propose an approach to rigorously evaluate the effect of LLM hallucinations during summarization. We demonstrate that \ul{summarizing pathology report text improves survival prediction and that LLM hallucinations do not impact survival predictions over their derived embeddings} (see Sections \ref{sec:report-results} and \ref{sec:hallucination-results}).
\end{itemize}

\begin{figure*}[t!]
\floatconts % label, caption, image
  {fig:overview}
  {\caption{Multimodal cancer modeling is vastly simplified in the age of foundation models. Simple ensembles of small, classical models over zero-shot FM embeddings improve pan-cancer survival risk stratification. Embedding methods are modular, allowing for simple experimentation and orchestration.}}
  {\includegraphics[width=0.9\linewidth]{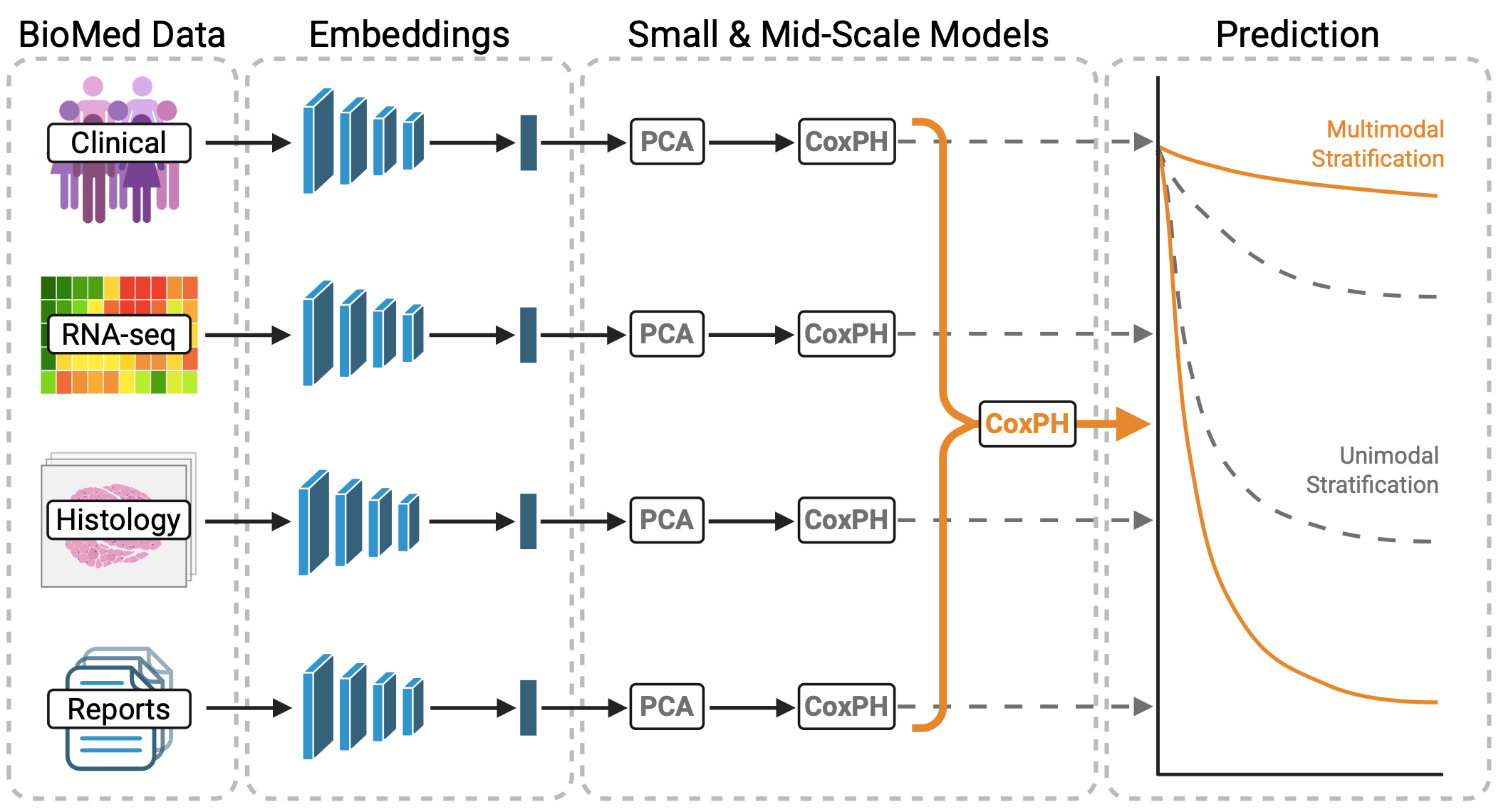}}
\end{figure*}

\section{Methods}
\subsection{Data and Experimental Setup}
We use TCGA patient cases which have available and valid survival data, pathology reports, tumor diagnostic slides, and tumor RNA-seq gene expression. For pathology reports, we use text extracted by \citet{kefeli2024tcga}. For computational efficiency, we additionally rely on precomputed UNI2 embeddings of diagnostic slides by Chen et al. \citep{chen2024towards}. Given these requirements, we first filter TCGA cases by the availability of pathology report, diagnostic slide, then RNA-seq data. We download the extracted text from Mendeley \citep{report-dataset}, precomputed UNI2 embeddings of diagnostic slides from HuggingFace \citep{uni2-dataset}, and RNA-seq data using the Genomic Data Commons (GDC) API \citep{heath2021nci}. We ensure that the RNA-seq and diagnostic slides are derived from tumor samples.

We then select cases with valid demographic and survival data, again downloaded using the GDC API. Specifically, we compute survival as the time between the patient's age at primary diagnosis and the patient's age at last followup or death. We additionally require non-missing patient sex. Given the final set of cases, we split our dataset for 5-fold cross-validation. We stratify by patient age, sex, race, ethnicity, mortality, and cancer type. To stratify by age, we discretize age into 20-year age bins, i.e. [0-20), [20-40), [40-60), [60-80), and 80\raisebox{0.25ex}{+}. For race and ethnicity, missing values are replaced with ``Not Reported''. We one-hot encode demographic features (binned age, sex, race, ethnicity) and cancer type (TCGA project) for stratification and downstream modeling, resulting in 17 and 32 features, respectively.

\subsection{Automated Report Summarization and Manual Correction}
\label{sec:summarization}

While the extracted pathology reports \citep{kefeli2024tcga} are a rich source of information, they present multiple challenges. As the source reports are scanned documents of varying quality, the OCR extraction process can result in typos and loss of structured text. Additionally, the reports are often long with repeated information or information potentially irrelevant for cancer prognosis (e.g. incidental findings on histology).

To address these challenges, we experiment with having an LLM summarize the pathology reports to help focus relevant information, correct typos, and reduce report length. We choose Llama-3.1-8B-Instruct \citep{grattafiori2024llama} as the summarization model for its strong instruction following capabilities. We use vLLM \citep{kwon2023efficient} for inference with set seed, greedy decoding, temperature 0, and max tokens 1,024. Our prompt is presented in Table \ref{tab:prompt}.

As we experiment with model generated summaries of pathology reports, we test the effect of model hallucinations \citep{huang2025survey} in the generated summaries and their downstream effects on survival modeling. To that end, we manually review model generated summaries for 40 randomly selected cases contained within a single test split. We randomly select these cases while preserving the overall prevalence of observed mortality. After review and manual correction (as needed) of the sampled summaries, we embed the corrected summaries using BioMistral and apply the survival model trained over BioMistral embeddings of summaries from the corresponding train split. We evaluate risk stratification over these sampled cases. Further details of our manual correction method are in Appendix \ref{apd:manual-correction}.

\begin{figure}[t!]
\floatconts % label, caption, image
    {fig:inclusion}
    {\caption{Eight thousand patient cases spanning 32 cancer types with survival data, pathology reports, diagnostic slides, and gene expression quantification.}}
    {\includegraphics[width=0.8\linewidth]{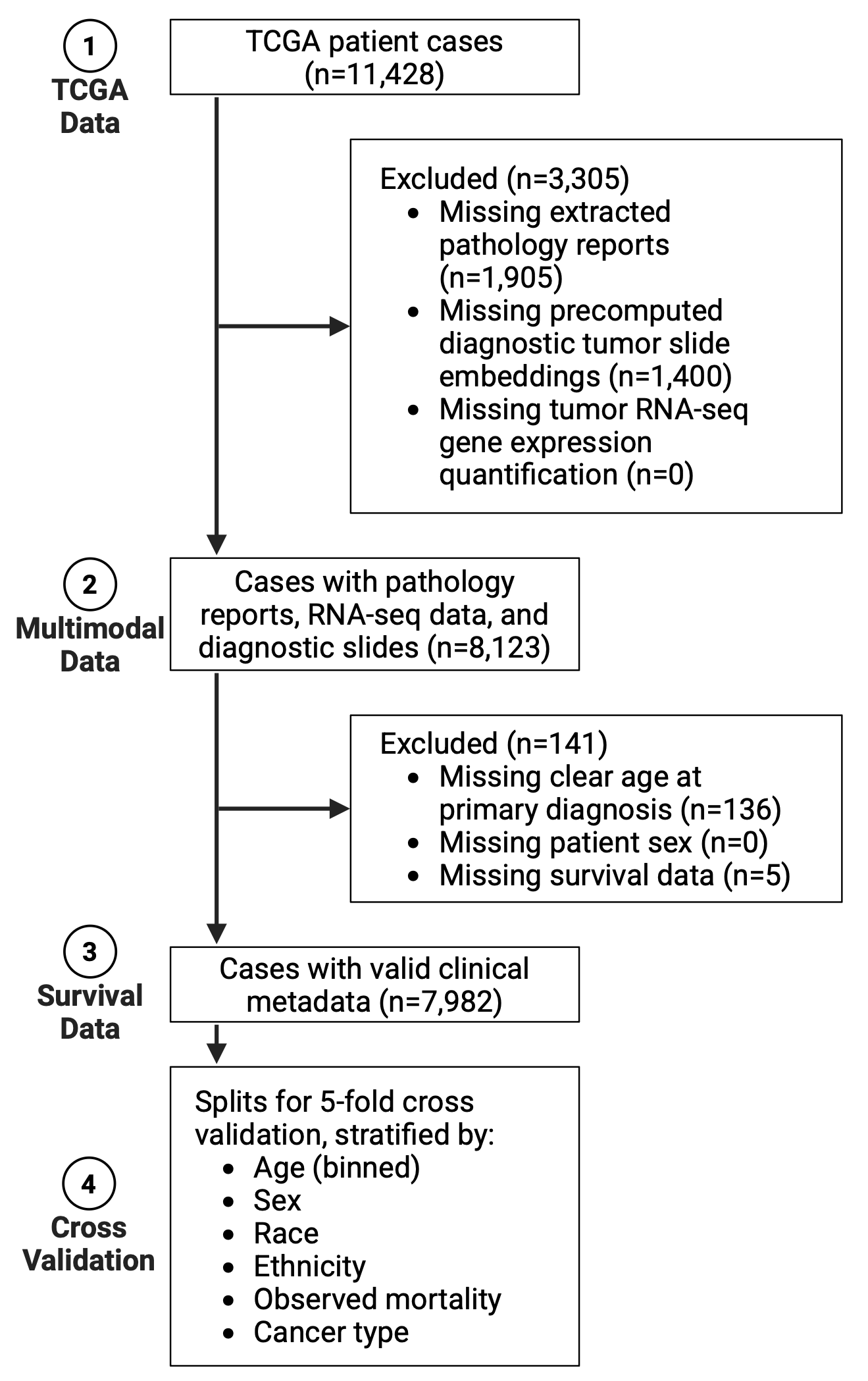}}
\end{figure}

\begin{table*}[t!]
\floatconts % label, caption, image
    {tab:unimodal}
    {\caption{Foundation model derived embeddings predict cancer survival. Average cross-validated C-index of CoxPH models trained over unimodal embeddings across varying PCA reductions. Demo: demographics; Canc: cancer type; Expr: BulkRNABert embedded RNA-seq; Hist: UNI2 embedded histology; Text: BioMistral embedded summarized pathology reports. *Not reduced with PCA.}}
    {\input{tables/unimodal}}
\end{table*}
\subsection{Foundation Model Embeddings}
We use several foundation models in our experiments. For all FMs, we use the models as-is with no further training or fine-tuning. \ul{To prevent data leakage, we ensure that the FMs we use have not been trained over TCGA}. Specifically, we experiment with UNI2-h \citep{chen2024towards} for diagnostic slides, BulkRNABert \citep{gelard2025bulkrnabert} or UCE \citep{rosen2023universal} for gene expression data, and BioMistral-7B \citep{labrak2024biomistral} or Mistral-7B-Instruct-v0.1 \citep{jiang2023mistral} for pathology reports. Further details on these models is presented in Appendix \ref{apd:fm-details}.

As survival is defined at the patient-level, we aggregate patient embeddings from multiple samples of the same modality into a single embedding per patient per modality. This is most relevant for RNA-seq and diagnostic slides as a patient may have multiple samples for these modalities. We use simple averaging of embeddings to derive our patient-level embeddings. Model embedding dimensionality shown in Table \ref{tab:dimensions}.

\subsection{Unimodal and Multimodal Survival Modeling and Evaluation}
\label{sec:surv-modeling}

Given patient-level, modality-specific embeddings, we adopt a simple pipeline for unimodal survival modeling. For a given train-test split, we z-score standardize all embeddings using per-feature mean and standard deviation derived from the training split. We next derive a Principal Component Analysis (PCA) \citep{pearson1901liii} dimensionality reduction over the standardized embeddings from the training split and apply to all standardized embeddings. We use dimensionality reduction to prevent overfitting and improve computational tractability (see Table \ref{tab:full-expr-hist} and Appendix \ref{apd:no-pca-embeddings} for results and discussion on using full embeddings). As PCA is an efficient transformation that is insensitive to hyperparameter choices (as in t-SNE or UMAP), we specifically use the implementation of PCA from sklearn \citep{pedregosa2011scikit} with a set seed. We experiment with varying PCA dimensions, doubling from 4 to 256. For demographic or cancer type modalities, features are one-hot encoded so we do not apply standardization or PCA transformation. We then fit a Cox Proportional Hazards (CoxPH) \citep{cox1972regression} model to the PCA reduced embeddings (or the one-hot encoded features for demographic or cancer type modalities) of the train split. We use the implementation of CoxPH models from sksurv \citep{polsterl2020scikit} with alpha = 0.1 for ridge regression penalty. Using the trained unimodal model, we predict risk scores for both the train and test splits for further modeling and evaluation.

We do late multimodal fusion using the predicted unimodal risk scores as input to the multimodal model. Given a set of modalities we aim to fuse, we concatenate the predicted risk scores from each modality as input features. We z-score standardize each feature using the mean and standard deviation derived over the training split. We fit a CoxPH model over the concatenated, standardized, unimodal risk scores. Such a model can be interpreted as modeling the unimodal risk (further discussed in Section \ref{sec:multimodal-results}). Using the trained multimodal model, we predict risk scores for the test split for evaluation. We repeat this multimodal fusion procedure over all combinations of demographic, cancer type, histology, expression, and text modalities. Notably, we do not consider alternate embedding methods as separate modalities; for example BioMistral embeddings of unsummarized and summarized reports are both text embeddings. Instead, we consider multimodal combinations with these alternate embedding strategies independently.

We evaluate the resulting models using 5-fold cross-validated concordance index (C-index) \citep{harrell1982evaluating}, mean cumulative/dynamic area under the receiver operating characteristic curve (mean AUC\textsuperscript{C,D}, where larger is better) \citep{lambert2016summary}, integrated Brier score (IBS, where smaller is better) \citep{brier1950verification}, or risk stratified Kaplan-Meier \citep{kaplan1958nonparametric} survival curves. Precise details on our evaluation method are provided in Appendix \ref{apd:evaluation}.

\section{Results}

% ===========================================================================
% ===========================================================================
\subsection{Multimodal data across 32 cancers}
% ===========================================================================
% ===========================================================================

We assemble a dataset of 7,982 patient cases for our study (Figure \ref{fig:inclusion}). These cases span 32 of the 33 cancer types in TCGA. We select cases that have available and valid data for survival analysis, preextracted pathology reports by Kefeli and Tatonelli \citep{kefeli2024tcga}, precomputed tumor diagnostic slide embeddings by Chen et al. \citep{chen2024towards}, and tumor RNA-seq gene expression data.

For our experiments, we split the dataset into 5 stratified cross-validation folds that are shared across all experiments. Descriptive features of our dataset and splits are provided in Tables \ref{tab:cohort} and \ref{tab:cancer-type}. While we stratify by cancer type and observed mortality, we do not stratify by survival duration. On average, overall survival time is comparable across splits, however per-cancer type subsets contained within each split may have too few observed mortalities for survival modeling (Table \ref{tab:cancer-type}). See Appendix \ref{apd:evaluation} for details on the limitations this presents for evaluating model performance within given cancer types.

\begin{table*}[t!]
\floatconts % label, caption, image
    {tab:best-multimodal}
    {\caption{Mixing survival models of clinical features with FM-based survival models improves survival prediction. For the best combination of each number of modality combinations, average cross-validated C-index of multimodal CoxPH models trained over predicted risk scores from unimodal CoxPH models across varying PCA reductions. Demo: demographics; Canc: cancer type; Expr: BulkRNABert embedded RNA-seq; Hist: UNI2 embedded histology; Text: BioMistral embedded summarized pathology reports. *Not reduced with PCA.}}
    {\input{tables/best-multimodal}}
\end{table*}

% ===========================================================================
% ===========================================================================
\subsection{Unimodal FM embeddings predict cancer survival}
\label{sec:unimodal-results}
% ===========================================================================
% ===========================================================================

We fit small, linear CoxPH survival models over unimodal data and attain a peak mean cross-validated C-index of approximately 0.75 over single modalities (Table \ref{tab:unimodal}). For categorical demographic features or cancer type, we do no dimensionality reduction. As compared to the mixture of models over both clinical tabular features (Canc-Demo C-index = 0.747, Table \ref{tab:best-multimodal}), CoxPH models fit over demographic features or cancer type alone achieve slightly lower C-index (0.630 and 0.737, respectively).

% Specifically, we fit and evaluate survival models over aggregated case-level embeddings for each modality.
For FM-derived embeddings (expression, histology, and text), we tested varying PCA sizes for dimensionality reduction. We observe a general trend of increasing C-index with increasing PCA dimensions, which plateaus around PCA to 256 dimensions. These findings are recapitulated using additional metrics of mean AUC\textsuperscript{C,D} and IBS (Tables \ref{tab:time-dependent-auroc} and \ref{tab:integrated-brier-score}, respectively). Based on this observation, we report further results using PCA transformation of these embeddings to 256 dimensions.

% We find that using BulkRNABert embeddings for tumor RNA-seq gene expression quantification, we achieve C-index = 0.750. We further find that survival modeling over UNI2 derived embeddings of tumor diagnostic histology slides results in C-index = 0.754. For pathology reports, using BioMistral derived embeddings of pathology reports summarized by Llama-3.1-8B-Instruct, we attain C-index = 0.752.

With the same unimodal survival models, we evaluate their survival prediction performance across individual cancer types (Tables \ref{tab:per-project-8} and \ref{tab:per-project}). We do not evaluate per-cancer survival based on cancer type as the input features would not differ within a cancer subset. We not only find that survival modeling performance of unimodal models varies across cancer types, we also observe that no single modality generally outperforms the other modalities. For example, among unimodal models, demographic-based modeling was most predictive of thyroid carcionma (THCA) survival (C-index = 0.885), while the same was true for text report-based modeling of endometrial carcinoma (UCEC) survival (C-index = 0.723).

% ===========================================================================
% ===========================================================================
\subsection{Multimodal fusion improves cancer modeling and survival prediction}
\label{sec:multimodal-results}
% ===========================================================================
% ===========================================================================

While unimodal models were able to achieve a peak C-index of approximately 0.75, we find that all multimodal fusion models are able to surpass unimodal results at their corresponding PCA reductions (Tables \ref{tab:best-multimodal} and \ref{tab:fm-multimodal}). We again observe a plateau of survival model performance at PCA = 256 and thus report results using this dimensionality reduction for embeddings. Importantly, each modality is independently modeled at a given PCA dimensionality before the predicted unimodal risk scores are subsequently used as input features to the multimodal fusion model (Section \ref{sec:surv-modeling}).

Using this simple, late multimodal fusion technique, we find that fusion of expression with histology data, expression with report data, and histology with report data result in mean cross-validated C-index = 0.774, 0.778, and 0.780, respectively (Table \ref{tab:fm-multimodal}). Fusion of all three of expression, histology, and reports results in the best embedding-based performance of C-index = 0.788. Survival modeling with these embedding-based modalities are not only additive, but also can be further enhanced with clinical features, such as demographics, achieving our greatest C-index = 0.795 (Table \ref{tab:best-multimodal}). The additive effect of multimodal fusion is further observed in both mean AUC\textsuperscript{C,D} and IBS (Tables \ref{tab:time-dependent-auroc} and \ref{tab:integrated-brier-score}, respectively). Importantly, these modalities are encoding information beyond simply cancer type, as any combination of modalities with cancer type improves upon the results of using cancer type alone (Table \ref{tab:combinations-w-cancer-type}).

\begin{table*}[htbp]
\floatconts % label, caption, image
    {tab:per-project-8}
    {\caption{Multimodal fusion improves survival prediction within the 8 most prevalent cancer types in the TCGA. Average cross-validated C-index of models using PCA to 256 dimensions for all foundation model derived embeddings. Multimodal model used fusion of all unimodal modalities.}}
    {\input{tables/per-project-8}}
\end{table*}

Additionally, as before, the ability to model survival within each cancer type further substantiates this claim, as the unimodal model based on cancer type cannot predict survival within a single cancer type. When examining multimodal model performance subset by cancer type in Tables \ref{tab:per-project-8} and \ref{tab:per-project}, we find that multimodal fusion consistently performs better than any other modality within a given cancer type for the most prevalent cancer types (see Appendix \ref{apd:evaluation} for additional details on evaluations within cancer types). This is in contrast to the unimodal models where the best unimodal model differs depending on the cancer.\\

\noindent\textbf{Interpretability.} Under our framework for multimodal fusion, a natural interpretation of the multimodal model arises. As the input to the multimodal CoxPH model are z-score normalized unimodal risk scores, the hazard ratios of the trained multimodal model can be interpreted relative to the standard deviations of the unimodal predicted risks. In other words, the hazard ratio for a given modality is the relative risk of death for a one standard deviation increase in that modality's unimodal predicted risk. Table \ref{tab:hazard-ratios} presents the hazard ratios of the 5-modality model (using unimodal models over PCA = 256 for high-dimensional embeddings). We find that the high-dimensional inputs confer greater relative risk, recapitulating our observations that these modalities are informative beyond tabular features.

% ===========================================================================
% ===========================================================================
\subsection{Pathology report summarization focuses cancer information}
\label{sec:report-results}
% ===========================================================================
% ===========================================================================

One of our key findings is that LLM summarization of pathology reports drastically improves survival prediction. Specifically, we zero-shot prompt Llama-3.1-8B-Instruct to summarize pathology reports with a focus on microscopic descriptions, test results, diagnoses, and clinical history. See Section \ref{sec:summarization} for more details on our summarization method.

Using Llama generated summaries, we embed the summarized reports using BioMistral. As compared with BioMistral embeddings of the original, unsummarized reports, we find that the summarized reports are able to better predict survival (C-index = 0.752 for summarized vs C-index = 0.694 for unsummarized). Using the predicted risk scores, we stratify the study cohort into low and high risk groups and observe that summarized reports result in enhanced risk stratification (Figure \ref{fig:summarization}).

Not only does summarization reduce the token length of the original report, it also corrects typographical errors that may result from text extraction from the source scanned PDFs of these reports. In addition, prompting the summarization to focus on certain aspects of the report may better extract information most relevant to survival prediction. With these effects in combination, we observe that embeddings of summarization outperforms embeddings of the original report across the 8 most prevalent cancer types (Table \ref{tab:per-project-text}).

% ===========================================================================
% ===========================================================================
\subsection{Hallucination correction does not impact cancer risk stratification}
\label{sec:hallucination-results}
% ===========================================================================
% ===========================================================================

While we empirically find that summarization improves survival prediction, hallucinations are an important consideration when using LLM generated text. To test whether factually incorrect information in the generated summaries impact downstream survival modeling, we manually correct generated summaries for these hallucinations.

To that end, we develop and share a lightweight tool to facilitate comparison of original and summarized reports. A screenshot of the tool is presented in Figure \ref{fig:comparison-tool}. Using this utility, we review 40 randomly sampled reports contained within a single cross-validation fold. We ultimately apply minor corrections to 24 of 40 reports. See Appendix \ref{apd:manual-correction} for additional details, such as methodology and categorization of manual corrections.

After manual correction, we re-embed the corrected summaries and do survival prediction using the same CoxPH model trained on uncorrected summarized embeddings. For the subset of corrected reports, we compare against the corresponding uncorrected summaries (Figure \ref{fig:hallucination}). We empirically find that risk stratification of the sampled subset does not change based on correction of hallucinations. This may be because the corrections on average are qualitatively minor compared to the majority of the summary (Table \ref{tab:correction-categories}). We hypothesize that the salient information for predicting cancer survival, such as cancer severity, extent of invasion or metastasis, remained unchanged, though further work is needed.

% ===========================================================================
% ===========================================================================
\subsection{Impact of domain specificity on cancer modeling and survival prediction}
% ===========================================================================
% ===========================================================================

Our final set of experiments were ablations to test the impact of foundation model domain specificity for downstream embedding-based modeling. For space, extended results are presented in Appendix \ref{apd:domain-results}. In brief, domain adaptation is particularly important for data modality specificity (BulkRNABert is notably better than UCE for RNA-seq data, Figure \ref{fig:expr-models}), however specificity to biomedical text (BioMistral vs Mistral) matters less than summarization (Figure \ref{fig:text-models}).

\begin{figure}[t!]
\floatconts % label, caption, image
    {fig:summarization}
    {\caption{Summarization of pathology reports improves survival model risk stratification over unimodal text embeddings. Embeddings derived with BioMistral. Reports summarized with Llama-3.1-8B-Instruct. Averaged risk stratification from 5-fold cross-validation.}}
    {\includegraphics[width=0.8\linewidth]{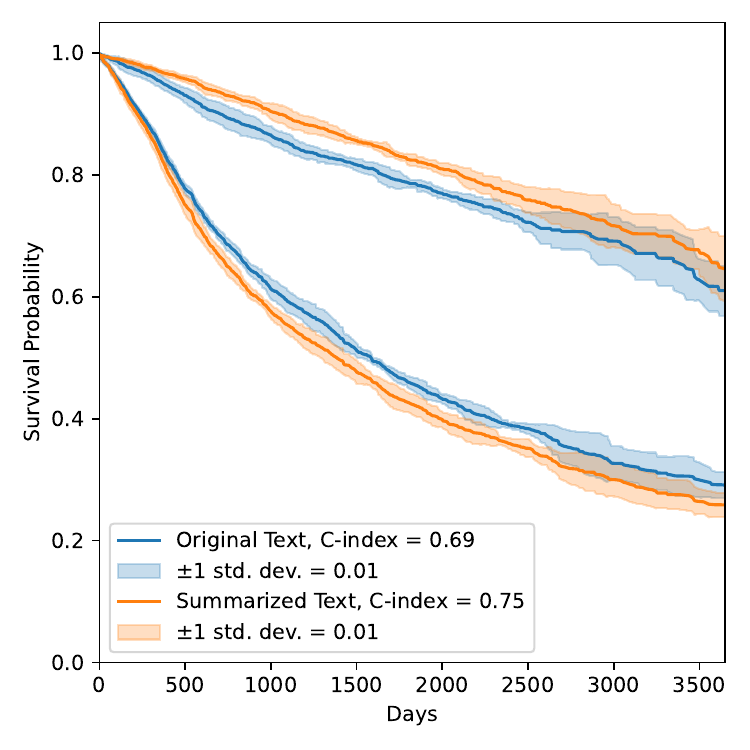}}
\end{figure}

\section{Discussion}
\label{sec:discussion}

Our study demonstrates the ease with which multimodal cancer modeling is accomplished in the age of foundation models. We show that zero-shot embeddings of unseen data can serve as the basis for prediction of cancer prognosis. We emphasize that our final survival models are combinations of classical CoxPH models with FM derived embeddings and five tabular, clinical features. \ul{The synergy between feature-rich embeddings and simple linear models enables prediction of complex biological tasks} \citep{ahlmann2024deep}. This powerful paradigm in modern deep learning is additionally beneficial in medical settings where data privacy is a chief concern and model tuning can lead to memorization of small datasets of protected patient data \citep{moor2023foundation, kaissis2020secure, khalid2023privacy, torkzadehmahani2022privacy}. In our results, we rely on FMs trained over deidentified and often public data; we do not train any deep learning models. It is important to note that this low-resource approach applies to settings that do not have the compute infrastructure for supervised training of bespoke deep learning models.

Importantly, we compare survival models over more complex modalities against survival models over basic clinical tabular features spanning four demographic features (age, sex, race, ethnicity) and cancer type. \ul{We observe that unimodal survival predictions over FM embeddings of expression, histology, or test data are no better than the combination of demographics and cancer type}. This is a critical baseline to contextualize the relative benefit of more complex methods over standard, classical models. Indeed, survival rates vary by cancer type as a complex interaction of biological and socioeconomic factors \citep{siegel2025cancer} and thus, such tabular features may simply be correlative. Additionally, survival prediction within a given cancer type cannot be predicted by the cancer type itself. For these reasons, unimodal models over FM embeddings may still have utility.

To that end, the results from our simple framework for multimodal survival modeling additionally show that \ul{extracted information from single modalities is additive and that multimodal fusion is able to achieve impressive prediction accuracy}. This additive effect may be explained by the varying unimodal results for per-cancer survival modeling, as we find that no single unimodal model excels across all cancer type subsets in TCGA. Conversely, our multimodal model excels across a majority of highly prevalent cancer types. Further, in sequential addition of modalities, we find that FM embeddings for these modalities are non-redundant (contrary to \citet{liu2025adaptive}) and encode information beyond simply cancer type.

Compared to other works investigating multimodal models, we propose a simple yet effective late fusion strategy. We ensemble the predicted outputs of unimodal models as input to the final multimodal model. This further \ul{modularizes the modeling of individual modalities (in addition to FM embeddings) and allows the multimodal model to be agnostic to unimodal embeddings}. This abstraction simplifies challenges of other fusion approaches. For example, ensuring smaller dimensionality features (such as demographics) are modeled comparably to higher dimensional features (such as expression embeddings) is accomplished implicitly in our framework. This modularity greatly facilitated our experimentation with varying FM embeddings per modality.

To the best of our knowledge, MUSK by \citet{xiang2025vision} is the only other work to have explored multimodal survival modeling with pathology reports. While they also utilize LLM summarization to condense their pathology reports, our work differs in key ways. First, unlike the MUSK authors, we do no training of our foundation models and evaluate zero-shot extracted embeddings. Yet their reported multimodal C-index of 0.747 is only comparable to our unimodal results, including naive cancer-type based prediction. Our multimodal fusion of histology and text resulted in C-index = 0.780. We additionally incorporate gene expression and clinical data, alongside histology and text modalities, achieving C-index = 0.795 and further surpassing their dual modality results. Experimentally, due to their language model length limitation, they cannot compare to unsummarized reports as we have. Lastly, their text summarization differs per cancer type, requiring specific, hand-written prompts by an oncologist; while this may focus the most relevant information for a given cancer type, \ul{we create a generalizable summarization strategy, applicable to any cancer type, that results in strong downstream predictions}.

We further extend our report summarization results to assess the impact that LLM hallucination has on downstream embedding and modeling. While we needed to manually correct 24 of 40 randomly selected summaries, many of these corrections qualitatively involved small details relative to the rest of the summary, e.g. correcting the number of lymph nodes negative for metastases. \ul{We found manual correction of LLM summarization hallucinations did not change the risk stratification} of the sampled subpopulation, compared to uncorrected summaries. While this was true in our study, we emphasize the importance of this kind of experiment, especially as researchers increasingly rely on LLM generated text.\\

\noindent\textbf{Limitations.} While we aim to do extensive experiments to validate our findings, we note key limitations. First, the small sample size of our hallucination correction experiment potentially limits its generalizability. Indeed, we call on other researchers to do such evaluation under their specific experimental settings for similar work relying on LLM text-generation as an intermediate step. Additionally, we leave variations on dimensionality reduction method and histology embedding models as future work. Lastly, while our work establishes a comprehensive baseline for cancer modeling of TCGA data, our results are currently limited to a single dataset. The primary factor for this limitation is finding data with all studied data modalities, particularly with pathology reports; it is ongoing work to do this external validation.

\section{Conclusion}
Overall, our study presents a modular framework for multimodal cancer modeling leveraging FM derived embeddings with classical models. We present a quantification of this approach applied to cancer survival prediction over the premier cancer data resource, The Cancer Genome Atlas. We show that multiple data modalities are additive using a simple late fusion technique, leaving the door open for expansion to additional modalities. We present novel results on pathology report-based survival prediction and the effects of LLM hallucinations in summarizing those reports. Altogether, these findings represent an important opportunity for the development, application, and evaluation of FMs towards cancer multi-omics and broader biomedicine.

\acks{This work was funded in part through the Advanced Research Projects Agency for Health (ARPA-H) under contract 75N92020D00021/5N92023F00002. The views and conclusions contained in this document are those of the authors and should not be interpreted as representing the oﬃcial policies, either expressed or implied, of the U.S. Government. S.S. is additionally supported by NIH training grant T32GM007281. Overview figure created in BioRender.}

\bibliography{references}

\clearpage
\appendix

\section{Manual Correction of Summaries}
\label{apd:manual-correction}

In our manual correction of generated summaries, we only change factually incorrect information based on information from the original report. We do not add extra information that was not already present in the summary. When the incorrect information cannot be corrected based on the original report, we delete the erroneous text. A salient example of this was when patient age was redacted in the original report. The resulting extracted text thus contained a fragment such as ``-year-old patient'', which the summarizing LLM interpreted to mean a 1-year-old patient. Manual verification of the case metadata revealed the patient to be in their 40s, however, because this data was impossible to derive from the original report, we remove the mention of the patient age in the corrected summary. All manual corrections for our experiment were done by a medical student who had completed two years of preclinical medical education. The sampled and corrected summaries are availabile in our GitHub repository.

For manual correction of summaries, we develop and share a lightweight, interactive tool for comparing unsummarized and summarized reports. The tool allows users to select sections of text in one text box which are automatically highlighted in the other text box. This enables users to quickly and interactively find corresponding information in large spans of text, thus facilitating verification of summaries using the source report. The tool is implemented in pure HTML and JavaScript and runs locally in a web browser, requiring no internet connection to use. We share the tool and a demo video of its use in our GitHub repository.

To better understand the kinds of corrections applied, we qualitatively categorize the 36 correction instances from 24 of 40 total reports in our experiment. We present the correction category counts in Table \ref{tab:correction-categories} and provide brief descriptions of these categories here. Descriptor or information mixup is when the summary incorrectly associates information described in multiple, potentially related parts of the report (e.g. swapping tumor vs benign tissue location). Lymph node information in particular was often incorrect, potentially due to the long spans of text that typically describes sentinel lymph node biopsies. A more benign type of error was copy-forward errors from the OCR text extraction process, e.g. NO vs N0 staging. Perhaps the most egregious hallucination was the inclusion of information not otherwise in the report, for example inferring the indication for the biopsy despite not being included in the original report. Gross description inclusion was considered an error given that we specifically prompted models to exclude gross description; the inclusion of the gross description thus represents conflation of gross and microscopic descriptors by the LLM. Relatedly, there were specific medical phrases used that the model would incorrectly interpret, such as “not otherwise specified”. A specific subtype of this that we identified was with misinterpretation of tumor staging (e.g. a placeholder value “X” to indicate missing or unreported was misinterpreted as the tumor grade; we verified there is no such grade for the given cancer type). Inclusion of patient age was also observed, despite all age being deidentified in the source reports. One summary was grammatically ambiguous. Finally, one correction should not have been included and was an error on the part of the manual corrector; the specific report with the error contained a different OCR error so overall the number of corrected reports remains 24.

\section{Foundation Model Details}
\label{apd:fm-details}

For diagnostic slides, we use UNI2-h \citep{chen2024towards} precomputed embeddings. Notably, as UNI2 is a tile/patch-level encoder, we aggregate tile embeddings to a slide-level embedding using simple averaging. For pathology reports or their summaries, we embed text using BioMistral-7B \citep{labrak2024biomistral} or its source model, Mistral-7B-Instruct-v0.1 \citep{jiang2023mistral}. We perform embedding inference for these LLMs using vLLM \citep{kwon2023efficient}. For gene expression data, we experiment with both BulkRNABert \citep{gelard2025bulkrnabert} and Universal Cell Embedding (UCE) \citep{rosen2023universal}. For BulkRNABert, to prevent data leakage, we specifically use the model checkpoint trained over GTEx \citep{lonsdale2013genotype} and ENCODE \citep{encode2012integrated} data. For UCE, we use the 33-layer model variant trained over data from CELLxGENE \citep{czi2025cz}. We make minor modifications to the code repositories for both of these models to enable installation as pip packages and inline data processing. Our modifications are contained in forks of these repositories linked from our main repo.

For the pathology report text modality, we choose BioMistral for its long context length and general biomedical domain adaptation, as opposed to other long-context LLMs adapted to more specific clinical domains \citep{yang2024clinicalmamba} unrelated to pathology reports. More specific pathology language models such as CONCH \citep{lu2024visual} or MUSK \citep{xiang2025vision} have been reported, however they are limited by extremely short contexts of 128 and 100 tokens, respectively. In our data, using the Mistral tokenizer (shared by BioMistral), the longest report is 8,184 tokens and the longest summarized report is 1,389 tokens. Despite BioMistral's domain adaptation with a context length of 2,048, the base Mistral model's context length of 8,196 and its sliding window attention \citep{beltagy2020longformer} enable a theoretical maximum context length that fits all of our pathology reports without truncation. Future work is needed to explore LLMs adapted to the pathology domain at the full context length.

\section{Evaluation of Survival Models}
\label{apd:evaluation}

Our primary evaluation metric for our survival models is the concordance index (C-index) \citep{harrell1982evaluating}. For each model, we report the average C-index derived over the test split from 5 cross-validation folds. For model performance by cancer type, we compute the C-index over only cases belonging to each cancer type; as our cross-validation folds are stratified by cancer type and mortality, the number of cases and observed deaths for a given cancer type across folds is approximately equivalent.

While we aim to understand our models' performance within each cancer type, a consideration with such evaluation is the limited number of potential samples within a given cancer type (see Table \ref{tab:cancer-type}), particularly when considering cross-validated results. This is an intrinsic limitation of TCGA and public cancer datasets in general, as some cancers are naturally less represented than others; for example, in the case of cholangiocarcinoma (CHOL), prior to data filtering, there are only 51 cases in TCGA. We thus present results in Table \ref{tab:per-project-8} over the 8 most prevalent cancer types and in Table \ref{tab:per-project} over as many cancer types as possible under our experimental setup. Specifically, for certain cancer types with very few observed deaths, as we rely on 5-fold cross-validation for results and due to the multifaceted stratification of our data splits, the observed mortality in specific splits may be extremely limited or is otherwise imperfectly stratified. When there are no observed deaths in a split (for the given cancer type) or when the observed death is after all other samples are right-censored, it is not possible to derive comparable pairs for computing concordance index. We thus can only present cross-validated results within 27 of the 32 total modeled cancer types. Additionally, while we can compute cross-validated, per-cancer performance for these 27 cancer types, it is important to consider the number of observed events used to compute these results. We thus report the average number of observed deaths and number of samples across test splits in our cross-validation.

To visualize the prognostic capability of our models, we plot averaged risk stratification curves. For a given test split, we binarize the predicted risk about the median into high and low risk groups. For both the low and high risk groups, we compute their Kaplan-Meier curves \citep{kaplan1958nonparametric}. We repeat this procedure for each cross-validation fold. To compute the average survival curve, we first impute the curves to a shared set of time points before averaging across folds. When visually appropriate, we include shaded regions about the average curves denoting one standard deviation at each time point.

To further validate our findings, we compute two additional evaluation metrics: mean cumulative/dynamic area under the receiver operating characteristic curve (mean AUC\textsuperscript{C,D}) \citep{lambert2016summary} and integrated Brier score (IBS) \citep{brier1950verification}. We specifically compute these metrics over the interval of 1 to 5 years. These boundaries correspond approximately to the 20th and 80th percentiles, respectively, of observation time points within our data.

The only exception to our cross-validated evaluation is in our experiment to manually correct hallucinations in model generated summaries. As cases for this experiment are contained within a single cross-validation fold, we do not report aggregate evaluations for this experiment.

\section{Domain specificity improves cancer modeling}
\label{apd:domain-results}

For bulk RNA-seq gene expression modality, we compare our default (BulkRNABert) against Universal Cell Embedding (UCE) \citep{rosen2023universal}, a foundation model trained over singe cell RNA-seq data. We hypothesized that a single-cell RNA-seq (scRNA-seq) FM may generalize to the mixed cellular identities of bulk RNA-seq data. While many scRNA-seq FMs exist \citep{szalata2024transformers}, we chose UCE specifically based on its reported strong performance at extracting multiscale embeddings of cellular biology \citep{rosen2023universal}. We find that while UCE derived embeddings do contain some prognostic signal (C-index = 0.637, PCA = 256) which stratifies risk (Figure \ref{fig:expr-models}), this is significantly lower compared to BulkRNABert derived embeddings (C-index = 0.753, PCA = 256) and even naive cancer type-based survival (C-index = 0.737, Table \ref{tab:unimodal}).

Furthermore, given the strong effects of summarization observed in Figure \ref{fig:summarization}, we test the impact of domain specificity for the text embedding. To isolate the effect of domain adaptation, we compare BioMistral derived embeddings against Mistral-7B-Instruct-v0.1 (Mistral) \citep{jiang2023mistral} derived embeddings. We compare BioMistral and Mistral embeddings for both original, unsummarized pathology reports and summarized reports (Figure \ref{fig:text-models}).

We find that for unsummarized reports, embedding FM domain adaptation has a small but appreciable effect on 5-fold cross-validated risk stratification, however it does not change survival prediction performance (Mistral C-index = 0.691 vs BioMistral C-index = 0.691, both PCA = 256). For summarized reports, both risk stratification and risk prediction are equivalent (Mistral C-index = 0.751 vs BioMistral C-index = 0.752). Rather, we recapitulate our pervious finding that summarization itself has a greater effect for survival modeling. Given their near equivalence and minor improvement in unsummarized report-based risk stratification, we use BioMistral as our default text embedding model.

\section{Using full embeddings is intractable or leads to overfitting}
\label{apd:no-pca-embeddings}

To test the necessity of dimensionality reduction in our framework, we compare our results to using the full FM-derived embeddings in each unimodal model. The modularity of our framework enables this experiment, however we note that using these full embeddings dramatically increases the computational complexity of the unimodal CoxPH models. When using the text embeddings of size 4096 (see all model embedding sizes in Table \ref{tab:dimensions}), a single CoxPH model with a maximum of 100 iterations (the default in the scikit-survival package) takes an impractical amount of time to train; we stopped the run after 2 hours. The histology embeddings of size 1536 required approximately 11 minutes per model on our hardware. For reference, when using PCA transformation, we can iterate through 7 different PCA dimensionality reduction experiments in 17 minutes on our compute infrastructure, representing 1,085 models in total (each reduction experiment runs 5-fold cross validation where each split has 31 models for all modality combinations i.e. the power set of 5 modalities). Given the full text embedding matrix of approximately 6k train-split samples by 4k covariates, fitting a Cox model over such an input becomes intractable, as the time complexity of training a Cox model is generally $\mathcal{O}(ndp^2)$, where $n$ is the number of samples, $d$ is the number of events, and $p$ is the number of covariates \citep{therneau2017using}. Due to these computational constraints, we are limited to only presenting the comparison to raw embeddings for histology and expression modalities.

Table \ref{tab:full-expr-hist} presents these results comparing the cross validated results for PCA=256 transformed version of embeddings vs the full expression or histology embeddings. We note that the full expression embeddings are size 256 and are thus its results are equivalent to those of the PCA=256 transformed expression embeddings. However, we observe substantial degradation of the unimodal histology model when using the full embedding input. We hypothesize that this is due to overfitting of the histology model to the training data given the high-dimensional histology embedding size of 1536. To verify this, we compare the cross-validated performance on the train vs test splits for both the full expression and full histology embedding models. Indeed, we find greater overfitting of the histology model to the training data when using no dimensionality reduction (Table \ref{tab:full-expr-hist-overfit}), providing empirical evidence for the inclusion of dimensionality reduction in our framework.

\clearpage
\onecolumn
\section{Extended Tables and Figures}
\label{apd:ext-tabs-figs}

\begin{table*}[htbp]
\floatconts % label, caption, image
    {tab:integrated-brier-score}
    {\caption{Average cross-validated integrated Brier score of CoxPH models across varying PCA reductions. Demo: demographics; Canc: cancer type; Expr: BulkRNABert embedded RNA-seq; Hist: UNI2 embedded histology; Text: BioMistral embedded summarized pathology reports; Multimodal: multimodal fusion of all modalities. *Not reduced with PCA.}}
    {\input{tables/integrated-brier-score}}
\end{table*}

\begin{table*}[htbp]
\floatconts % label, caption, image
    {tab:time-dependent-auroc}
    {\caption{Average cross-validated cumulative/dynamic area under the ROC curve of CoxPH models across varying PCA reductions. Demo: demographics; Canc: cancer type; Expr: BulkRNABert embedded RNA-seq; Hist: UNI2 embedded histology; Text: BioMistral embedded summarized pathology reports; Multimodal: multimodal fusion of all modalities. *Not reduced with PCA.}}
    {\input{tables/time-dependent-auroc}}
\end{table*}

\begin{table*}[htbp]
\floatconts % label, caption, image
    {tab:fm-multimodal}
    {\caption{Late, multimodal fusion of unimodal, FM-based survival models improves survival prediction. For modality combinations with only FM-derived embeddings, average cross-validated C-index of multimodal CoxPH models trained over predicted risk scores from unimodal CoxPH models across varying PCA reductions. Expr: BulkRNABert embedded RNA-seq; Hist: UNI2 embedded histology; Text: BioMistral embedded summarized pathology reports.}}
    {\input{tables/fm-multimodal}}
\end{table*}

\begin{table*}[htbp]
\floatconts % label, caption, image
    {tab:combinations-w-cancer-type}
    {\caption{Data modalities encode information beyond cancer type. Addition of other data modalities with cancer type consistently outperforms using cancer type alone. Canc: cancer type; Demo: demographics; Expr: BulkRNABert embedded RNA-seq; Hist: UNI2 embedded histology; Text: BioMistral embedded summarized pathology reports. Average cross-validated results reported using PCA to 256 dimensions for all foundation model derived embeddings.}}
    {\input{tables/combinations-w-cancer-type}}
\end{table*}

\begin{table*}[htbp]
\floatconts % label, caption, image
    {tab:hazard-ratios}
    {\caption{Hazard ratios of full multimodal fusion model. All modalities confer greater relative risk compared to cancer type. Canc: cancer type; Demo: demographics; Expr: BulkRNABert embedded RNA-seq; Hist: UNI2 embedded histology; Text: BioMistral embedded summarized pathology reports. Results reported using PCA to 256 dimensions for all foundation model derived embeddings.}}
    {\input{tables/hazard-ratios}}
\end{table*}

\begin{table*}[htbp]
\floatconts % label, caption, image
    {tab:per-project-text}
    {\caption{Summarization of pathology reports improves pan-cancer survival prediction. Average cross-validated C-index of models evaluated on subsets of the top 8 most prevalent cancer types in TCGA. All text embedded using BioMistral. Summarized text generated using Llama-3.1-8B-Instruct. Multimodal model incorporates unimodal summarized text model. Results reported using PCA to 256 dimensions for all foundation model derived embeddings.}}
    {\input{tables/per-project-text}}
\end{table*}

\begin{table*}[htbp]
\floatconts % label, caption, image
    {tab:prompt}
    {\caption{Pathology report summarization prompt.}}
    {\input{tables/prompt}}
\end{table*}

\begin{table*}[htbp]
\floatconts % label, caption, image
    {tab:correction-categories}
    {\caption{Categorization of manual corrections for hallucination experiment. Out of 40 sampled reports, 24 reports required manual corrections, totaling 36 specific instances of corrections. These corrections are generally qualitatively minor, though specific error types are more severe than others.}}
    {\input{tables/correction-categories}}
\end{table*}

\begin{figure}[htbp]
\floatconts % label, caption, image
    {fig:hallucination}
    {\caption{Manual correction of summarized pathology report hallucinations does not impact survival model risk stratification. Embeddings derived with BioMistral. Reports summarized with Llama-3.1-8B-Instruct. Risk stratification from N=40 randomly sampled cases contained within a single test split while preserving observed mortality prevalence.}}
    {\includegraphics[width=0.4\linewidth]{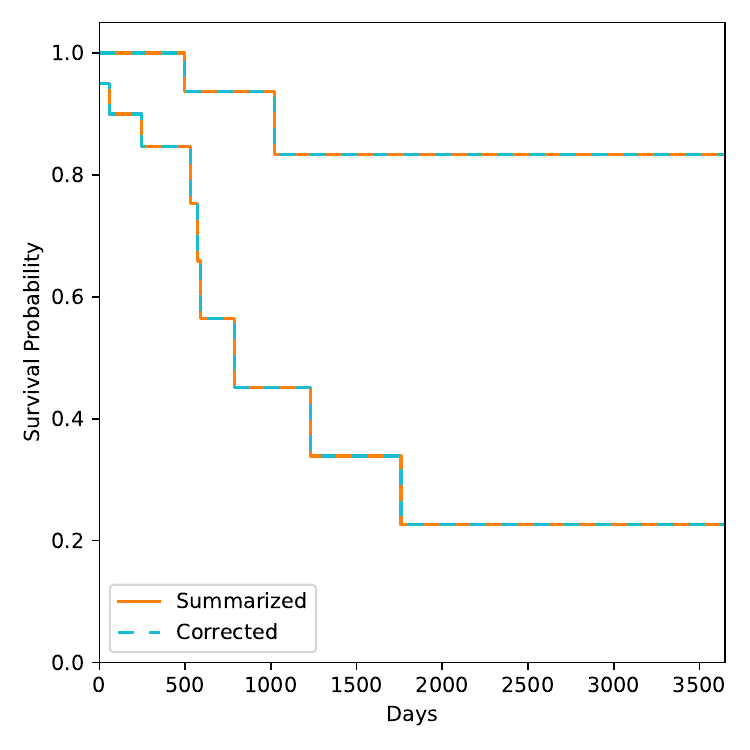}}
\end{figure}

\begin{figure}[htbp]
\floatconts % label, caption, image
    {fig:expr-models}
    {\caption{Modality specificity of gene expression embedding model improves survival model risk stratification over unimodal gene expression embeddings. Bulk RNA-seq data embedded with either BulkRNABert or UCE, a single-cell RNA-seq model. Averaged risk stratification from 5-fold cross-validation.}}
    {\includegraphics[width=0.4\linewidth]{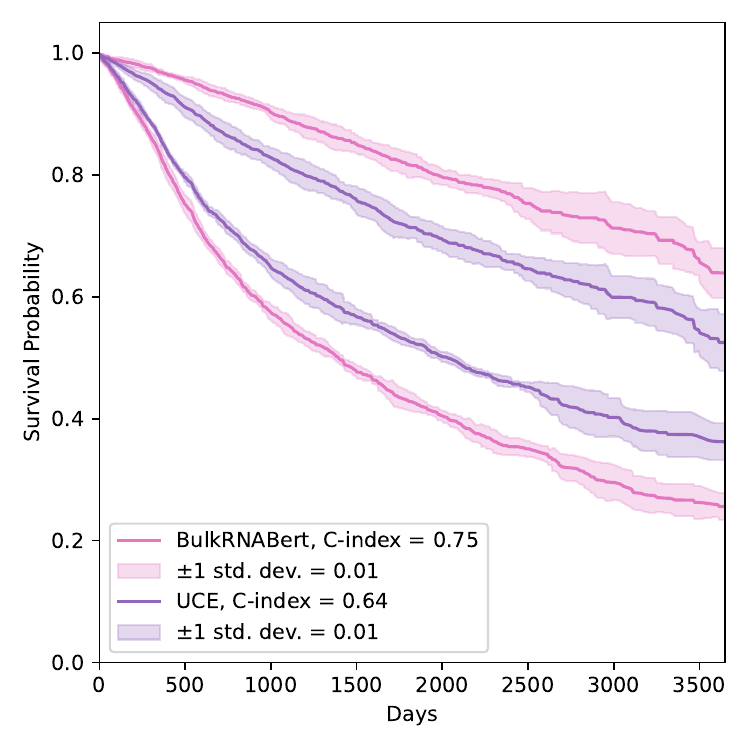}}
\end{figure}

\begin{figure}[htbp]
\floatconts % label, caption, image
    {fig:text-models}
    {\caption{Domain adaptation of text embedding model improves survival model risk stratification over unimodal text embeddings. Embeddings derived with either BioMistral or Mistral-7B-Instruct-v0.1. Reports summarized with Llama-3.1-8B-Instruct. Averaged risk stratification from 5-fold cross-validation.}}
    {\includegraphics[width=0.4\linewidth]{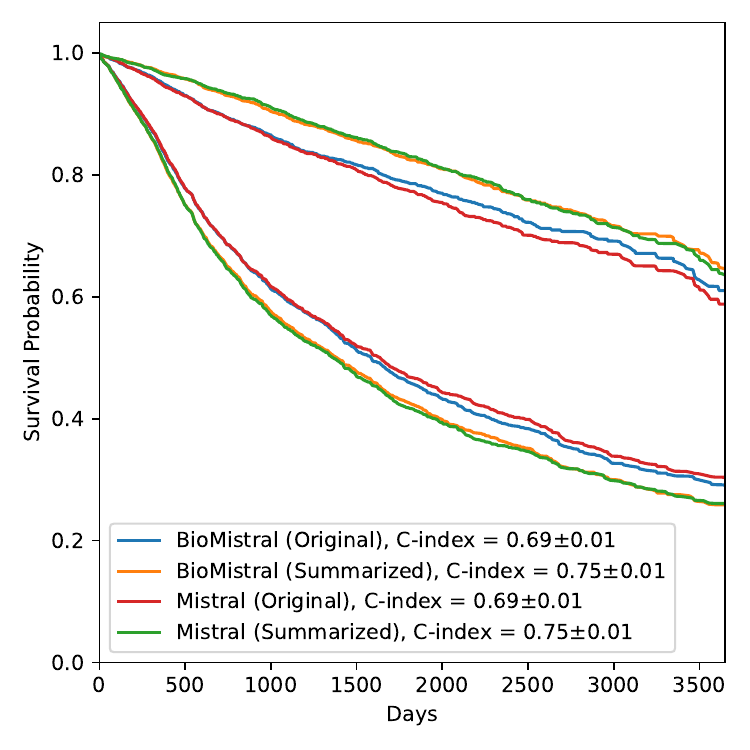}}
\end{figure}

\begin{table*}[htbp]
\floatconts % label, caption, image
    {tab:dimensions}
    {\caption{Foundation model embedding dimensionality before PCA reduction. Demo: demographics; Canc: cancer type; Expr: RNA-Seq gene expression; Hist: histology, Text: pathology reports.}}
    {\input{tables/dimensions}}
\end{table*}

\begin{table*}[htbp]
\floatconts % label, caption, image
    {tab:full-expr-hist}
    {\caption{Average cross-validated C-index of CoxPH models comparing PCA=256 with no dimensionality reduction. Demo: demographics; Canc: cancer type; Expr: BulkRNABert embedded RNA-seq; Hist: UNI2 embedded histology; Text: BioMistral embedded summarized pathology reports; Multimodal: multimodal fusion of all modalities. *Text embedding still PCA reduced to 256d.}}
    {\input{tables/full-expr-hist}}
\end{table*}

\begin{table*}[htbp]
\floatconts % label, caption, image
    {tab:full-expr-hist-overfit}
    {\caption{Average cross-validated C-index of CoxPH models comparing train vs test set overfitting/generalization when using full expression or histology embeddings. Expr: BulkRNABert embedded RNA-seq; Hist: UNI2 embedded histology. *Not reduced with PCA.}}
    {\input{tables/full-expr-hist-overfit}}
\end{table*}

\begin{sidewaystable*}[htbp]
\floatconts % label, caption, image
    {tab:cohort}
    {\caption{Survival prediction using eight thousand patient cases spanning 32 cancer types from TCGA. All cases have clinical, RNA-seq gene expression, diagnostic histology slides, and pathology reports. Patients split for 5-fold cross-validation, stratified by age bins, sex, race, ethnicity, observed mortality, and cancer type. AA: African American; AIAN: American Indian or Alaska Native; NHPI: Native Hawaiian or Pacific Islander.}}
    {\input{tables/cohort}}
\end{sidewaystable*}

\begin{table*}[htbp]
\floatconts % label, caption, image
    {tab:cancer-type}
    {\caption{Eight thousand patient cases span 32 cancer types from TCGA. Cancer types stratified across 5-fold cross-validation splits.}}
    {\input{tables/cancer-type}}
\end{table*}

\begin{table*}[htbp]
\floatconts % label, caption, image
    {tab:per-project}
    {\caption{Multimodal fusion improves survival prediction within given cancer types. Average cross-validated C-index of models evaluated on 27 cancer types in TCGA for which we could compute cross-validated results. Multimodal model used fusion of all unimodal modalities. Results reported using PCA to 256 dimensions for all foundation model derived embeddings. Average cross-validation test fold characteristics provided to demonstrate vanishingly small sample sizes.}}
    {\input{tables/per-project}}
\end{table*}

\begin{figure*}[htbp]
\floatconts % label, caption, image
    {fig:comparison-tool}
    {\caption{Manual review of generated summaries is facilitated by our simple comparison utility. Users dynamically highlight text between reports and can edit as needed. This is more powerful than traditional diff-checkers as summarized reports often fix typos or contain semantically similar text that is not an exact string match. The tool is implemented in pure HTML/JavaScript and does not require an internet connection to use.}}
    {\includegraphics[width=\linewidth]{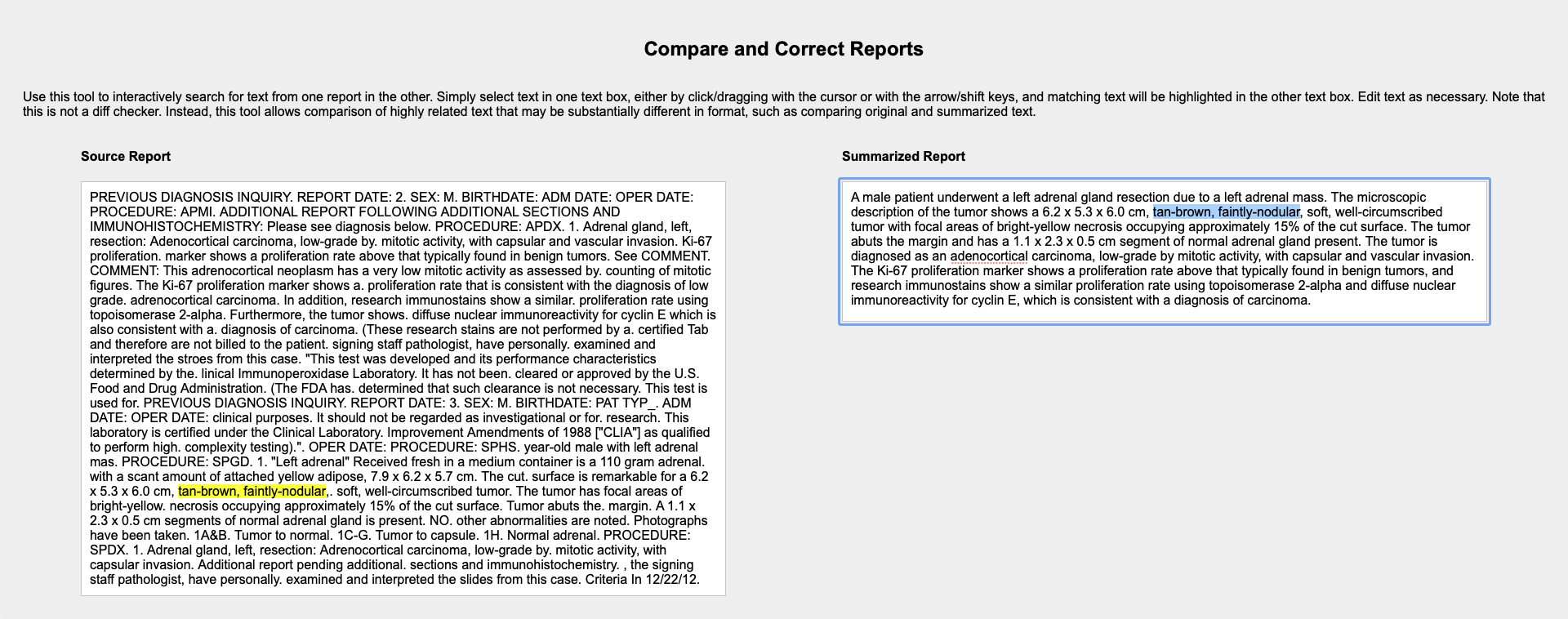}}
\end{figure*}

\end{document}

%% file: tables/unimodal.tex
\begin{tabular}{lrrrrrrrr}
\toprule
\multirow{2}{*}{Modality} & \multicolumn{8}{c}{PCA Dimension} \\
\cmidrule{2-9}
 & 4 & 8 & 16 & 32 & 64 & 128 & 256 & N/A\\
\midrule
Demo* & - & - & - & - & - & - & - & 0.630 \\
Canc* & - & - & - & - & - & - & - & 0.737\\
Expr & 0.626 & 0.650 & 0.702 & 0.742 & 0.749 & 0.753 & 0.750 & - \\
Hist & 0.596 & 0.631 & 0.669 & 0.714 & 0.733 & 0.748 & 0.754 & - \\
Text & 0.565 & 0.610 & 0.690 & 0.725 & 0.745 & 0.751 & 0.752 & - \\
\bottomrule
\end{tabular}

%% file: tables/best-multimodal.tex
\begin{tabular}{lrrrrrrrr}
\toprule
\multirow{2}{*}{Modality} & \multicolumn{8}{c}{PCA Dimension} \\
\cmidrule{2-9}
 & 4 & 8 & 16 & 32 & 64 & 128 & 256 & N/A \\
\midrule
Canc*-Demo* & - & - & - & - & - & - & - & 0.747 \\
Hist-Text & 0.606 & 0.662 & 0.705 & 0.746 & 0.767 & 0.777 & 0.780 & - \\
Expr-Hist-Text & 0.644 & 0.682 & 0.727 & 0.765 & 0.780 & 0.786 & 0.788 & - \\
Demo*-Expr-Hist-Text & 0.680 & 0.717 & 0.748 & 0.776 & 0.788 & 0.794 & 0.795 & - \\
Canc*-Demo*-Expr-Hist-Text & 0.749 & 0.752 & 0.760 & 0.777 & 0.788 & 0.793 & 0.793 & - \\
\bottomrule
\end{tabular}

%% file: tables/per-project-8.tex
\begin{tabular}{lrrrrrrrrr}
\toprule
\multirow{2}{*}{Modality} & \multicolumn{8}{c}{Cancer Type (TCGA Project)} \\
\cmidrule{2-9}
 & BRCA & KIRC & UCEC & THCA & LGG & HNSC & LUSC & LUAD \\
\midrule
Demographics & 0.637 & 0.600 & 0.593 & 0.885 & 0.727 & 0.525 & 0.542 & 0.530 \\
RNA-seq & 0.573 & 0.673 & 0.643 & 0.619 & 0.790 & 0.560 & 0.579 & 0.620 \\
Histology & 0.648 & 0.685 & 0.703 & 0.749 & 0.753 & 0.636 & 0.598 & 0.617 \\
Text & 0.641 & 0.698 & 0.723 & 0.619 & 0.716 & 0.582 & 0.556 & 0.645 \\
\midrule
Multimodal & 0.720 & 0.752 & 0.758 & 0.787 & 0.840 & 0.650 & 0.626 & 0.687 \\
\bottomrule
\end{tabular}

%% file: tables/integrated-brier-score.tex
\begin{tabular}{lrrrrrrrr}
\toprule
\multirow{2}{*}{Modality} & \multicolumn{8}{c}{PCA Dimension} \\
\cmidrule{2-9}
 & 4 & 8 & 16 & 32 & 64 & 128 & 256 & N/A\\
\midrule
Demo* & - & - & - & - & - & - & - & 0.179 \\
Canc* & - & - & - & - & - & - & - & 0.155 \\
Expr & 0.182 & 0.177 & 0.167 & 0.156 & 0.154 & 0.153 & 0.154 & - \\
Hist & 0.182 & 0.178 & 0.173 & 0.165 & 0.160 & 0.155 & 0.153 & - \\
Text & 0.185 & 0.181 & 0.166 & 0.159 & 0.155 & 0.152 & 0.153 & - \\
\midrule
Multimodal & 0.152 & 0.151 & 0.150 & 0.145 & 0.142 & 0.140 & 0.140 & - \\
\bottomrule
\end{tabular}

%% file: tables/time-dependent-auroc.tex
\begin{tabular}{lrrrrrrrr}
\toprule
\multirow{2}{*}{Modality} & \multicolumn{8}{c}{PCA Dimension} \\
\cmidrule{2-9}
 & 4 & 8 & 16 & 32 & 64 & 128 & 256 & N/A\\
\midrule
Demo* & - & - & - & - & - & - & - & 0.650 \\
Canc* & - & - & - & - & - & - & - & 0.764 \\
Expr & 0.642 & 0.674 & 0.728 & 0.771 & 0.779 & 0.783 & 0.778 & - \\
Hist & 0.620 & 0.647 & 0.690 & 0.737 & 0.760 & 0.779 & 0.785 & - \\
Text & 0.587 & 0.634 & 0.721 & 0.754 & 0.773 & 0.779 & 0.779 & - \\
\midrule
Multimodal & 0.779 & 0.783 & 0.790 & 0.809 & 0.820 & 0.826 & 0.825 & - \\
\bottomrule
\end{tabular}

%% file: tables/fm-multimodal.tex
\begin{tabular}{lrrrrrrr}
\toprule
\multirow{2}{*}{Modality} & \multicolumn{7}{c}{PCA Dimension} \\
\cmidrule{2-8}
 & 4 & 8 & 16 & 32 & 64 & 128 & 256 \\
\midrule
Expr-Hist & 0.638 & 0.665 & 0.714 & 0.753 & 0.765 & 0.771 & 0.774 \\
Expr-Text & 0.636 & 0.671 & 0.724 & 0.762 & 0.774 & 0.779 & 0.778 \\
Hist-Text & 0.606 & 0.662 & 0.705 & 0.746 & 0.767 & 0.777 & 0.780 \\
Expr-Hist-Text & 0.644 & 0.682 & 0.727 & 0.765 & 0.780 & 0.786 & 0.788 \\
\bottomrule
\end{tabular}

%% file: tables/combinations-w-cancer-type.tex
\begin{tabular}{lr}
\toprule
Combination & C-index \\
\midrule
Canc & 0.737 \\
Canc-Demo & 0.747 \\
Canc-Expr & 0.758 \\
Canc-Hist & 0.765 \\
Canc-Text & 0.761 \\
Canc-Demo-Expr & 0.767 \\
Canc-Demo-Hist & 0.773 \\
Canc-Demo-Text & 0.771 \\
Canc-Expr-Hist & 0.774 \\
Canc-Expr-Text & 0.776 \\
Canc-Hist-Text & 0.779 \\
Canc-Demo-Expr-Hist & 0.782 \\
Canc-Demo-Expr-Text & 0.784 \\
Canc-Demo-Hist-Text & 0.787 \\
Canc-Expr-Hist-Text & 0.786 \\
Canc-Demo-Expr-Hist-Text & 0.793 \\
\bottomrule
\end{tabular}

%% file: tables/hazard-ratios.tex
\begin{tabular}{lrrrrr|r}
\toprule
Split & 0 & 1 & 2 & 3 & 4 & Mean \\
\midrule
Demo & 1.301 & 1.329 & 1.340 & 1.290 & 1.356 & 1.323 \\
Canc & 0.646 & 0.640 & 0.637 & 0.661 & 0.625 & 0.642 \\
Expr & 1.899 & 1.960 & 1.998 & 1.959 & 1.991 & 1.961 \\
Hist & 2.067 & 2.049 & 2.052 & 2.030 & 2.072 & 2.054 \\
Text & 2.183 & 2.209 & 2.193 & 2.203 & 2.176 & 2.193 \\
\bottomrule
\end{tabular}

%% file: tables/per-project-text.tex
\begin{tabular}{lrrrrrrrr}
\toprule
\multirow{2}{*}{Modality} & \multicolumn{8}{c}{Cancer Type (TCGA Project)} \\
\cmidrule{2-9}
 & BRCA & KIRC & UCEC & THCA & LGG & HNSC & LUSC & LUAD \\
\midrule
Original & 0.612 & 0.645 & 0.575 & 0.593 & 0.630 & 0.574 & 0.540 & 0.593 \\
Summarized & 0.641 & 0.698 & 0.723 & 0.619 & 0.716 & 0.582 & 0.556 & 0.645 \\
\midrule
Multimodal & 0.720 & 0.752 & 0.758 & 0.787 & 0.840 & 0.650 & 0.626 & 0.687 \\
\bottomrule
\end{tabular}

%% file: tables/prompt.tex
\begin{tabular}{lll}
\toprule
Line & Role & Message \\
\midrule
1 & System & You are a helpful assistant for digital pathology.\\
\midrule
2 & System & \makecell[l]{Instructions:\\
Extract and repeat the results of the following pathology report\\in a single paragraph.\\
Focus on test results, diagnoses and clinical history.\\
Include results of the microscopic description.\\
Omit the gross or macroscopic description.\\
Do not acknowledge this prompt.\\
Do not give additional comments after your final answer.}\\
\midrule
3 & User & (Pathology Report) \\
\bottomrule
\end{tabular}

%% file: tables/correction-categories.tex
\begin{tabular}{lr}
\toprule
Correction Type & N \\
\midrule
descriptor/information mixup & 6 \\
incorrect lymph node information & 5 \\
copy-forward OCR error & 5 \\
made up information not otherwise in report & 4 \\
inclusion of gross description & 4 \\
misunderstood medical phrase & 4 \\
incorrect tumor staging & 3 \\
inclusion of patient age & 3 \\
grammar clarification & 1 \\
manual correction error & 1 \\
\bottomrule
\end{tabular}

%% file: tables/dimensions.tex
\begin{tabular}{llrc}
\toprule
Modality & Model & Embedding Size & Is Subsequently PCA Reduced? \\
\midrule
Demo & N/A & 17 & \xmark \\
Canc & N/A & 32 & \xmark \\
Hist & UNI2 & 1536 & \cmark \\
Text & BioMistral & 4096 & \cmark \\
Text & Mistral & 4096 & \cmark \\
Expr & BulkRNABert & 256 & \cmark \\
Expr & UCE & 1280 & \cmark \\
\bottomrule
\end{tabular}

%% file: tables/full-expr-hist.tex
\begin{tabular}{lrr}
\toprule
 & PCA=256 & No PCA \\
\midrule
Demo & - & 0.630 \\
Canc & - & 0.737 \\
Expr & 0.750 & 0.750 \\
Hist & 0.754 & 0.700 \\
Text & 0.752 & - \\
Multimodal & 0.793 & 0.729* \\
\bottomrule
\end{tabular}

%% file: tables/full-expr-hist-overfit.tex
\begin{tabular}{lrr}
\toprule
 & Train & Test \\
\midrule
Expr* & 0.785 & 0.750 \\
Hist* & 0.875 & 0.700 \\
\bottomrule
\end{tabular}

%% file: tables/cohort.tex
\begin{tabular}{llllllll}
\toprule
 &  &  & \multicolumn{5}{c}{Cross-Validation Fold} \\
\cline{4-8}
 &  & Overall & 0 & 1 & 2 & 3 & 4 \\
\midrule
n &  & 7982 & 1597 & 1597 & 1596 & 1596 & 1596 \\
\cline{1-8}
Age, mean (SD) &  & 59.8 (14.4) & 59.7 (14.4) & 59.9 (14.2) & 59.9 (14.5) & 59.9 (14.3) & 59.8 (14.4) \\
\cline{1-8}
\multirow[t]{2}{*}{Sex, n (\%)} & Female & 4192 (52.5) & 828 (51.8) & 842 (52.7) & 845 (52.9) & 836 (52.4) & 841 (52.7) \\
 & Male & 3790 (47.5) & 769 (48.2) & 755 (47.3) & 751 (47.1) & 760 (47.6) & 755 (47.3) \\
\cline{1-8}
\multirow[t]{7}{*}{Race, n (\%)} & White & 6019 (75.4) & 1212 (75.9) & 1203 (75.3) & 1207 (75.6) & 1203 (75.4) & 1194 (74.8) \\
 & Black or AA & 794 (9.9) & 154 (9.6) & 152 (9.5) & 165 (10.3) & 160 (10.0) & 163 (10.2) \\
 & Asian & 378 (4.7) & 72 (4.5) & 78 (4.9) & 77 (4.8) & 78 (4.9) & 73 (4.6) \\
 & AIAN & 20 (0.3) & 3 (0.2) & 6 (0.4) & 2 (0.1) & 3 (0.2) & 6 (0.4) \\
 & NHPI & 10 (0.1) & 3 (0.2) & 2 (0.1) & 1 (0.1) & 1 (0.1) & 3 (0.2) \\
 & Unknown & 123 (1.5) & 24 (1.5) & 28 (1.8) & 24 (1.5) & 21 (1.3) & 26 (1.6) \\
 & Not Reported & 638 (8.0) & 129 (8.1) & 128 (8.0) & 120 (7.5) & 130 (8.1) & 131 (8.2) \\
\cline{1-8}
\multirow[t]{4}{*}{Ethnicity, n (\%)} & Not Hispanic/Latino & 5983 (75.0) & 1197 (75.0) & 1197 (75.0) & 1197 (75.0) & 1192 (74.7) & 1200 (75.2) \\
 & Hispanic/Latino & 309 (3.9) & 62 (3.9) & 56 (3.5) & 59 (3.7) & 68 (4.3) & 64 (4.0) \\
 & Unknown & 172 (2.2) & 37 (2.3) & 39 (2.4) & 40 (2.5) & 31 (1.9) & 25 (1.6) \\
 & Not Reported & 1518 (19.0) & 301 (18.8) & 305 (19.1) & 300 (18.8) & 305 (19.1) & 307 (19.2) \\
\cline{1-8}
\multirow[t]{2}{*}{Mortality, n (\%)} & Alive & 5790 (72.5) & 1166 (73.0) & 1150 (72.0) & 1152 (72.2) & 1165 (73.0) & 1157 (72.5) \\
 & Dead & 2192 (27.5) & 431 (27.0) & 447 (28.0) & 444 (27.8) & 431 (27.0) & 439 (27.5) \\
\cline{1-8}
\multicolumn{2}{l}{Survival Time in Days, mean (SD)} & 1026.6 (964.2) & 1048.0 (978.4) & 1029.3 (969.7) & 1013.8 (956.0) & 1021.0 (961.9) & 1020.8 (955.5) \\
\bottomrule
\end{tabular}

%% file: tables/cancer-type.tex
\begin{tabular}{llllllll}
\toprule
 &  & & \multicolumn{5}{c}{Cross-Validation Fold} \\
\cline{4-8}
 &  & Overall & 0 & 1 & 2 & 3 & 4 \\
\midrule
 & n & 7982 & 1597 & 1597 & 1596 & 1596 & 1596 \\
\cline{1-8}
\multirow{32}{*}{\rotatebox[origin=c]{90}{Cancer Type, n (\%)}} & ACC & 53 (0.7) & 11 (0.7) & 9 (0.6) & 11 (0.7) & 11 (0.7) & 11 (0.7) \\
 & BLCA & 343 (4.3) & 67 (4.2) & 69 (4.3) & 71 (4.4) & 67 (4.2) & 69 (4.3) \\
 & BRCA & 982 (12.3) & 197 (12.3) & 194 (12.1) & 197 (12.3) & 198 (12.4) & 196 (12.3) \\
 & CESC & 248 (3.1) & 49 (3.1) & 47 (2.9) & 50 (3.1) & 50 (3.1) & 52 (3.3) \\
 & CHOL & 33 (0.4) & 8 (0.5) & 6 (0.4) & 5 (0.3) & 6 (0.4) & 8 (0.5) \\
 & COAD & 391 (4.9) & 78 (4.9) & 77 (4.8) & 77 (4.8) & 81 (5.1) & 78 (4.9) \\
 & DLBC & 43 (0.5) & 9 (0.6) & 9 (0.6) & 8 (0.5) & 9 (0.6) & 8 (0.5) \\
 & ESCA & 116 (1.5) & 23 (1.4) & 26 (1.6) & 23 (1.4) & 23 (1.4) & 21 (1.3) \\
 & GBM & 142 (1.8) & 29 (1.8) & 29 (1.8) & 28 (1.8) & 28 (1.8) & 28 (1.8) \\
 & HNSC & 433 (5.4) & 89 (5.6) & 88 (5.5) & 87 (5.5) & 83 (5.2) & 86 (5.4) \\
 & KICH & 65 (0.8) & 13 (0.8) & 13 (0.8) & 13 (0.8) & 13 (0.8) & 13 (0.8) \\
 & KIRC & 497 (6.2) & 97 (6.1) & 100 (6.3) & 102 (6.4) & 99 (6.2) & 99 (6.2) \\
 & KIRP & 237 (3.0) & 48 (3.0) & 48 (3.0) & 46 (2.9) & 48 (3.0) & 47 (2.9) \\
 & LGG & 442 (5.5) & 88 (5.5) & 90 (5.6) & 89 (5.6) & 86 (5.4) & 89 (5.6) \\
 & LIHC & 319 (4.0) & 62 (3.9) & 63 (3.9) & 63 (3.9) & 67 (4.2) & 64 (4.0) \\
 & LUAD & 411 (5.1) & 84 (5.3) & 83 (5.2) & 82 (5.1) & 80 (5.0) & 82 (5.1) \\
 & LUSC & 418 (5.2) & 82 (5.1) & 82 (5.1) & 85 (5.3) & 85 (5.3) & 84 (5.3) \\
 & MESO & 66 (0.8) & 15 (0.9) & 13 (0.8) & 12 (0.8) & 12 (0.8) & 14 (0.9) \\
 & OV & 42 (0.5) & 7 (0.4) & 10 (0.6) & 9 (0.6) & 7 (0.4) & 9 (0.6) \\
 & PAAD & 168 (2.1) & 34 (2.1) & 32 (2.0) & 34 (2.1) & 34 (2.1) & 34 (2.1) \\
 & PCPG & 171 (2.1) & 34 (2.1) & 37 (2.3) & 34 (2.1) & 34 (2.1) & 32 (2.0) \\
 & PRAD & 326 (4.1) & 67 (4.2) & 66 (4.1) & 66 (4.1) & 63 (3.9) & 64 (4.0) \\
 & READ & 144 (1.8) & 29 (1.8) & 28 (1.8) & 29 (1.8) & 30 (1.9) & 28 (1.8) \\
 & SARC & 240 (3.0) & 49 (3.1) & 50 (3.1) & 47 (2.9) & 47 (2.9) & 47 (2.9) \\
 & SKCM & 94 (1.2) & 19 (1.2) & 19 (1.2) & 17 (1.1) & 19 (1.2) & 20 (1.3) \\
 & STAD & 269 (3.4) & 52 (3.3) & 54 (3.4) & 55 (3.4) & 54 (3.4) & 54 (3.4) \\
 & TGCT & 87 (1.1) & 19 (1.2) & 16 (1.0) & 15 (0.9) & 18 (1.1) & 19 (1.2) \\
 & THCA & 483 (6.1) & 96 (6.0) & 96 (6.0) & 97 (6.1) & 98 (6.1) & 96 (6.0) \\
 & THYM & 109 (1.4) & 20 (1.3) & 21 (1.3) & 23 (1.4) & 24 (1.5) & 21 (1.3) \\
 & UCEC & 494 (6.2) & 99 (6.2) & 100 (6.3) & 97 (6.1) & 99 (6.2) & 99 (6.2) \\
 & UCS & 52 (0.7) & 11 (0.7) & 10 (0.6) & 10 (0.6) & 10 (0.6) & 11 (0.7) \\
 & UVM & 64 (0.8) & 12 (0.8) & 12 (0.8) & 14 (0.9) & 13 (0.8) & 13 (0.8) \\
\bottomrule
\end{tabular}

%% file: tables/per-project.tex
{
\setlength{\tabcolsep}{4.5pt}
\begin{tabular}{lrrrrrrrrr}
\toprule
\multirow{2}{*}{} & \multicolumn{9}{c}{Cancer Type (TCGA Project)} \\
\cmidrule{2-10}
 & BRCA & KIRC & UCEC & THCA & LGG & HNSC & LUSC & LUAD & COAD \\
\midrule
Mean Test Size (N) & 196.4 & 99.4 & 98.8 & 96.6 & 88.4 & 86.6 & 83.6 & 82.2 & 78.2 \\
Mean Test Mortality (N) & 28.0 & 33.4 & 15.8 & 3.2 & 20.0 & 40.6 & 37.0 & 29.8 & 15.8 \\
\midrule
Demographics & 0.637 & 0.600 & 0.593 & 0.885 & 0.727 & 0.525 & 0.542 & 0.530 & 0.563 \\
RNA-seq & 0.573 & 0.673 & 0.643 & 0.619 & 0.790 & 0.560 & 0.579 & 0.620 & 0.604 \\
Histology & 0.648 & 0.685 & 0.703 & 0.749 & 0.753 & 0.636 & 0.598 & 0.617 & 0.629 \\
Text & 0.641 & 0.698 & 0.723 & 0.619 & 0.716 & 0.582 & 0.556 & 0.645 & 0.704 \\
\midrule
Multimodal & 0.720 & 0.752 & 0.758 & 0.787 & 0.840 & 0.650 & 0.626 & 0.687 & 0.718 \\
\midrule
\midrule
\multirow{2}{*}{} & \multicolumn{9}{c}{Cancer Type (TCGA Project)} \\
\cmidrule{2-10}
 & BLCA & LIHC & STAD & CESC & SARC & KIRP & PAAD & READ & GBM \\
\midrule
Mean Test Size (N) & 68.6 & 63.8 & 53.8 & 49.6 & 48.0 & 47.4 & 33.6 & 28.8 & 28.4 \\
Mean Test Mortality (N) & 32.2 & 20.4 & 22.6 & 12.0 & 17.8 & 6.6 & 17.4 & 4.0 & 23.4 \\
\midrule
Demographics & 0.584 & 0.558 & 0.525 & 0.602 & 0.571 & 0.498 & 0.545 & 0.706 & 0.596 \\
RNA-seq & 0.594 & 0.557 & 0.579 & 0.670 & 0.620 & 0.798 & 0.634 & 0.538 & 0.506 \\
Histology & 0.604 & 0.590 & 0.591 & 0.619 & 0.590 & 0.719 & 0.610 & 0.587 & 0.589 \\
Text & 0.576 & 0.549 & 0.644 & 0.603 & 0.659 & 0.788 & 0.616 & 0.642 & 0.498 \\
\midrule
Multimodal & 0.663 & 0.608 & 0.676 & 0.704 & 0.673 & 0.846 & 0.673 & 0.664 & 0.570 \\
\midrule
\midrule
\multirow{2}{*}{} & \multicolumn{9}{c}{Cancer Type (TCGA Project)} \\
\cmidrule{2-10}
 & ESCA & SKCM & MESO & KICH & UVM & ACC & UCS & OV & CHOL \\
\midrule
Mean Test Size (N) & 23.2 & 18.8 & 13.2 & 13.0 & 12.8 & 10.6 & 10.4 & 8.4 & 6.6 \\
Mean Test Mortality (N) & 12.2 & 5.2 & 11.2 & 1.8 & 4.2 & 3.8 & 6.8 & 4.2 & 3.4 \\
\midrule
Demographics & 0.579 & 0.513 & 0.503 & 0.615 & 0.684 & 0.591 & 0.631 & 0.518 & 0.492 \\
RNA-seq & 0.447 & 0.617 & 0.611 & 0.937 & 0.652 & 0.826 & 0.495 & 0.653 & 0.466 \\
Histology & 0.596 & 0.495 & 0.643 & 0.805 & 0.675 & 0.713 & 0.440 & 0.530 & 0.437 \\
Text & 0.553 & 0.619 & 0.514 & 0.882 & 0.530 & 0.754 & 0.654 & 0.498 & 0.557 \\
\midrule
Multimodal & 0.576 & 0.590 & 0.665 & 0.937 & 0.739 & 0.825 & 0.628 & 0.588 & 0.493 \\
\bottomrule
\end{tabular}
}